\documentclass[letterpaper]{article} 
\usepackage{aaai2026}  
\usepackage{times}  
\usepackage{helvet}  
\usepackage{courier}  
\usepackage[hyphens]{url}  
\usepackage{graphicx} 
\usepackage{natbib}  
\usepackage{caption} 
\usepackage{algorithm}
\usepackage{algorithmic}

\usepackage{newfloat}
\usepackage{listings}
\DeclareCaptionStyle{ruled}{labelfont=normalfont,labelsep=colon,strut=off} 
\floatstyle{ruled}
\newfloat{listing}{tb}{lst}{}
\floatname{listing}{Listing}

\usepackage{subcaption}
\usepackage{amsmath}
\usepackage{amssymb}
\usepackage{booktabs}
\newcommand{\sysname}{{SASR{} }}
\title{Step-wise Adaptive Integration of Supervised Fine-tuning and Reinforcement Learning for Task-Specific LLMs}
\author{
    Jack Chen\textsuperscript{\rm 1,3}\thanks{These authors contributed equally to this work.},
    Fazhong Liu\textsuperscript{\rm 2}\footnotemark[1],
    Naruto Liu\textsuperscript{\rm 1,3},
    Yuhan Luo\textsuperscript{\rm 2},
    Erqu Qin\textsuperscript{\rm 1,3},
    Harry Zheng\textsuperscript{\rm 1,3},\\
    Tian Dong\textsuperscript{\rm 2},
    Haojin Zhu\textsuperscript{\rm 2},
    Yan Meng\textsuperscript{\rm 2}\thanks{Corresponding author.},
    Xiao Wang\textsuperscript{\rm 1,3}\footnotemark[2]
}
\affiliations{
    \textsuperscript{\rm 1}Shanghai Goku Technologies Limited\\
    \textsuperscript{\rm 2}Shanghai Jiao Tong University\\
    \textsuperscript{\rm 3}Shanghai AllMind Artificial Intelligence Technology Co., Ltd.\\
    
    \textsuperscript{\rm 1,3}\{chenzhi2, liutian2, qinxiao2, zhengziwei2, wangxiao\}@gokudata.com\\
    \textsuperscript{\rm 2}\{liufazhong, gilerure, tian.dong, zhu-hj, yan\_meng\}@sjtu.edu.cn
}

\begin{document}

\maketitle

\begin{abstract}

Large language models (LLMs) excel in mathematical reasoning and logical problem-solving. Current mainstream paradigms rely on supervised fine-tuning (SFT) or reinforcement learning (RL), yet each possesses distinct strengths: SFT enables the model to internalize fixed patterns, whereas RL empowers it to actively explore and acquire transferable decision-making strategies under reward guidance, thereby enhancing generalization. Recent state-of-the-art methods therefore advocate hybrid schemes, but static switching yields poor cross-task generalization . Inspired by the \textit{curriculum learning-quiz} mechanism in human reasoning cultivation, we present \sysname, a \underline{s}tep-wise \underline{a}daptive hybrid training framework that theoretically unifies \underline{S}FT and \underline{R}L and dynamically balances them throughout optimization. \sysname begins with an SFT warm-up to establish basic reasoning skills, then employs an adaptive algorithm that monitors gradient norms and divergence from the initial distribution to seamlessly integrate SFT with the online RL method GRPO. By tracking the model’s training status and sequentially adjusting the process, \sysname ensures a smooth transition between paradigms, preserving core reasoning while exploring diverse paths. Experimental results demonstrate that \sysname outperforms SFT, RL, and static hybrid training methods.

\end{abstract}

\section{Introduction}

Large language models (LLMs) have demonstrated remarkable capabilities in complex reasoning tasks, including mathematical problem-solving~\cite{Romera-Paredes2024, ahn-etal-2024-large}, symbolic manipulation~\cite{pan-etal-2023-logic, 10002218} and multi-step logical inference~\cite{wang2024qimprovingmultistepreasoning, NEURIPS2022_9d560961, wang2023selfconsistency}. These advances are largely driven by sophisticated training paradigms that combine supervised fine-tuning (SFT) with reinforcement learning (RL). SFT supplies the model with high-quality, step-by-step reasoning demonstrations—often in the form of chain-of-thought (CoT) annotations—thereby instilling structured problem-solving strategies. RL, in turn, refines these capabilities through reward-driven optimization, aligning outputs with human preferences or task-specific objectives. This hybrid approach underpins state-of-the-art models such as GPT-4~\cite{openai2024gpt4technicalreport}, DeepSeek~\cite{deepseekai2025deepseekr1incentivizingreasoningcapability} and Claude~\cite{anthropic_claude_2025}, achieving unprecedented benchmark performance.

However, as the demand grows to train LLMs for specialized tasks that lack large-scale, high-quality datasets yet still require strong reasoning, the prevailing paradigms reveal critical shortcomings. Primitive training schemes—namely pure SFT or pure RL—each pose distinct challenges. SFT heavily relies on meticulously curated CoT annotations and verified ground-truth answers, rendering it susceptible to overfitting~\cite{li2024entropic, fu2024dispersethenmergepushinglimitsinstruction, gekhman-etal-2024-fine}. RL, conversely, suffers from reward hacking~\cite{hu2025open} and mode collapse~\cite{liu2025understanding}, which can erode the model’s acquired reasoning skills. To mitigate these issues, researchers have proposed auxiliary techniques such as entropy bonuses~\cite{luo2025deepscaler}, curriculum learning~\cite{team2025kimi} and PTX loss~\cite{ouyang2022training}. Yet these methods remain confined to a single paradigm and therefore fail to address its fundamental weaknesses.
Consequently, recent work explores hybrid paradigms that integrate SFT and RL, and such approaches have demonstrated superior performance~\cite{guo2025deepseek}. For example, DeepSeek employs a two-stage schedule that first performs SFT and then switches to RL. Nevertheless, the efficacy of such static schedules remains questionable when high-quality data are scarce, and their generalization across diverse tasks has yet to be rigorously established.

\begin{figure}[h]
\centering
\includegraphics[width=\linewidth]{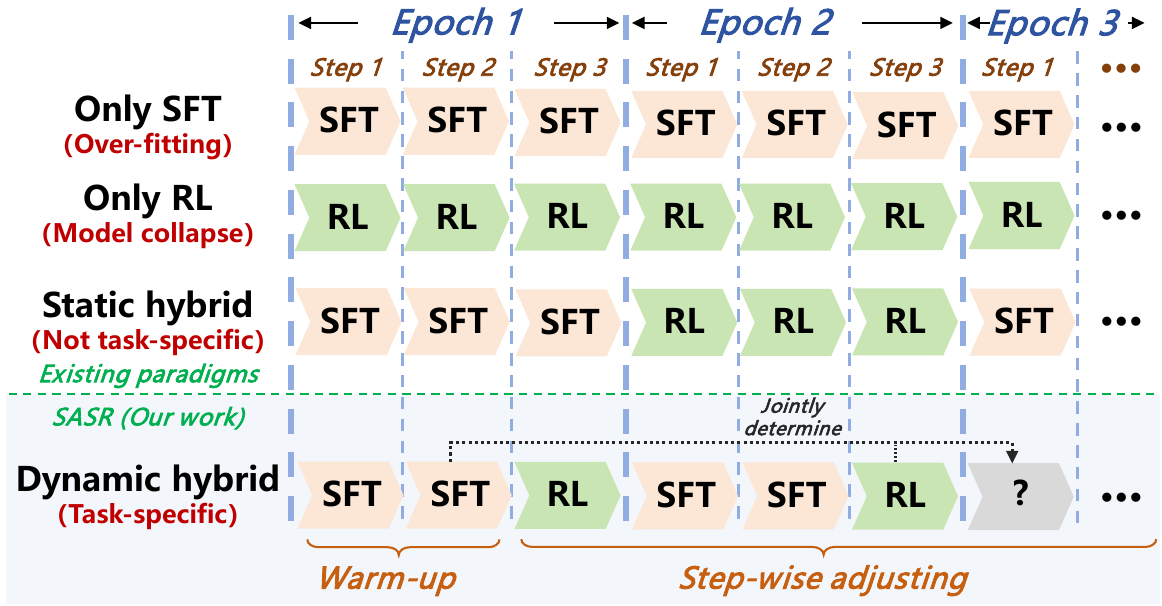}
\caption{Comparison of different training paradigms.}
\label{fig:stragety}

\end{figure}

To address these challenges, we propose \sysname, an adaptive hybrid training method inspired by the human process of structured learning followed by practice. In \sysname, SFT serves as guided study using reference materials, whereas RL acts as a quiz-like reinforcement that fosters generalization. \sysname unifies these two stages and adaptively modulates their proportions according to the model’s training dynamics.

Just as students must first study solved examples before tackling novel problems, \sysname begins with an SFT warm-up to establish basic reasoning capabilities. Subsequently, it seamlessly blends SFT with the online RL method GRPO. This stage mirrors students who, after reviewing worked examples, solve new problems to enhance generalization. Abruptly removing all reference materials after the warm-up risks pushing the model away from sound reasoning patterns. \sysname therefore dynamically adjusts the SFT/GRPO ratio throughout training, guided by the model’s evolving state. Concretely, at every training step, the ratio is updated by comparing current gradient norms with those recorded during the warm-up phase.

By continuously monitoring gradient magnitudes and learning trends, \sysname achieves a smooth transition between paradigms, balancing the retention of basic skills with the exploration of diverse reasoning paths. In contrast, existing hybrid methods~\cite{deepseekai2025deepseekr1incentivizingreasoningcapability, luong2024reftreasoningreinforcedfinetuning, havrilla2024teaching} rely either on static schedules—fixing the number of SFT and RL epochs a priori—or on hard switches that abruptly change paradigms without any transition phase.
We identify and formalize a long-overlooked tension—the \textit{forget-stagnation trade-off}. Upon transitioning from SFT to RL, RL’s exploratory nature shifts the policy away from the initial distribution, resulting in catastrophic forgetting of basic reasoning skills. Conversely, excessive reliance on SFT to maintain stability leads to stagnation: the model overfits to known patterns and generalizes poorly to novel tasks. This trade-off is exacerbated in static or hard-switching hybrids, which either cannot adapt to evolving learning states or omit any transitional phase altogether.

To systematically mitigate this trade-off, \sysname introduces a gradient-norm-based adaptive mechanism that recasts the forget-stagnation dilemma as a monitorable optimization objective, enabling progressive capability gains while safeguarding core reasoning skills.

We conduct extensive experiments on two base models—DeepSeek and Qwen—across three standard datasets: GSM8K, MATH and Knight-and-Knives (KK). Tasks span mathematical calculation and logic-based question answering. Results show that \sysname significantly outperforms SFT, RL and static hybrid baselines. On mathematical reasoning tasks, \sysname improves absolute accuracy by 12.45\% over SFT and 15.30\% over RL; on the more challenging MATH and KK datasets, it surpasses static hybrids by 8.0\% on average.

Our contributions are summarized as follows:
\begin{itemize}
\item We propose \sysname, the first adaptive dynamic training framework that theoretically unifies SFT and RL, demonstrating the superiority of smooth, data-driven hybrid training.
\item Drawing on the human \textit{curriculum learning-quiz} process, we design a dynamic switching indicator that leverages the relationship between warm-up gradient norms and current training states to resolve the forget-stagnation trade-off.
\item We validate \sysname on diverse mathematical-reasoning and logical-inference tasks, providing comprehensive empirical evidence of its effectiveness.
\end{itemize}


\section{Related Work}
\label{gen_inst}

\subsection{Mathematical Reasoning \& Logical Inference Solving}

Enhancing the mathematical problem solving and logical reasoning capabilities of LLMs has become a key focus in recent research. Various methods such as Chain-of-Thought (CoT) prompting~\cite{NEURIPS2022_9d560961} with its variants tree-of-thought~\cite{yao2023treethoughtsdeliberateproblem} and graph-of-thought~\cite{10.1609/aaai.v38i16.29720}, and self-consistency mechanisms~\cite{wang2023selfconsistency}, have demonstrated promising results on tasks like math word problems and arithmetic reasoning by encouraging coherent intermediate steps. However, these methods also face limitations. For instance, early-stage LLMs may generate unreliable explanations when performing few-shot textual reasoning~\cite{10.5555/3600270.3602472}. Additionally, while combining self-supervised learning with reward-model-based reinforcement learning can guide LLMs in solving mathematical problems~\cite{uesato2022solvingmathwordproblems}, concerns remain regarding reward hacking and the difficulty in capturing fine-grained logical inferences~\cite{deepseekai2025deepseekr1incentivizingreasoningcapability}.

\subsection{Training Paradigms} 

Supervised Fine-Tuning (SFT) is a foundational technique for adapting pre-trained language models to downstream tasks by training on high-quality demonstration data. Specifically, LLMs update their parameters by minimizing the discrepancy between predictions and ground-truth labels via gradient-based optimization. Due to its high efficiency and low cost, SFT has been widely adopted for fine-tuning LLMs in specialized areas such as mathematical reasoning~\cite{cobbe2021trainingverifierssolvemath, hendrycksmath2021, yuan2023scaling, gou2024tora}. However, SFT alone may struggle with open-ended tasks where optimal responses are less well-defined~~\cite{inverseRLignment}. 

In 2022, OpenAI introduce ChatGPT~\cite{ChatGPT}, a large language dialogue model that catalyzed the development and adoption of a Reinforcement Learning from Human Feedback (RLHF)~\cite{10.5555/3600270.3602281}, as a novel training paradigm in the field of LLM training. Advanced Reinforcement Learning (RL) methods such as Proximal Policy Optimization (PPO)~\cite{schulman2017proximalpolicyoptimizationalgorithms}, and Group Relative Policy Optimization (GRPO)~\cite{shao2024deepseekmathpushinglimitsmathematical} have since been integrated into refinement of LLMs. GRPO, which aims to enhance policy optimization by leveraging group-relative advantages, can notably improve mathematical reasoning capabilities with less memory consumption~\cite{shao2024deepseekmathpushinglimitsmathematical}. However, it can still encounter challenges like reward hacking~\cite{wen2024language} and pattern collapse~\cite{shao2024deepseekmathpushinglimitsmathematical}.

Given the limitations of utilizing only SFT or RL approaches, recent work explores hybrid approaches that combine both paradigms to enhance both instruction-following and reasoning. DeepSeek-R1~\cite{deepseekai2025deepseekr1incentivizingreasoningcapability} exemplifies this trend by employing a novel training framework that combines SFT and GRPO. Similarly, Reinforced Fine-Tuning (ReFT)~\cite{luong2024reftreasoningreinforcedfinetuning} demonstrates that RL-augmented fine-tuning can outperform pure SFT. These approaches highlight the potential of combining supervised learning with RL to unlock advanced reasoning in LLMs. However, the static hard-switching training scheme has considerable room for improvement in terms of dynamic task adaptation and progressive capability transfer.

\section{SASR: Step-wise Adaptive Integration of SFT and RL}
\label{Method}

\begin{figure*}[t]
\centering
    \begin{subfigure}[b]{0.53\textwidth}
    \includegraphics[width=\textwidth]{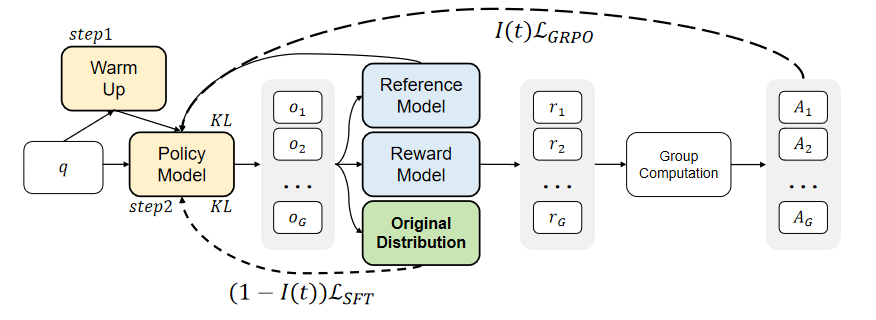}
    \caption{Visualizations of \sysname's architecture: warm-up  in step1, and  the training state is monitored through the condition function $I(t)$ to adaptively adjust the training paradigm in step2.}
    \label{fig:arch}
    \end{subfigure}
    \hspace{0.05\textwidth}
    \begin{subfigure}[b]{0.4\textwidth}
    \includegraphics[width=\textwidth]{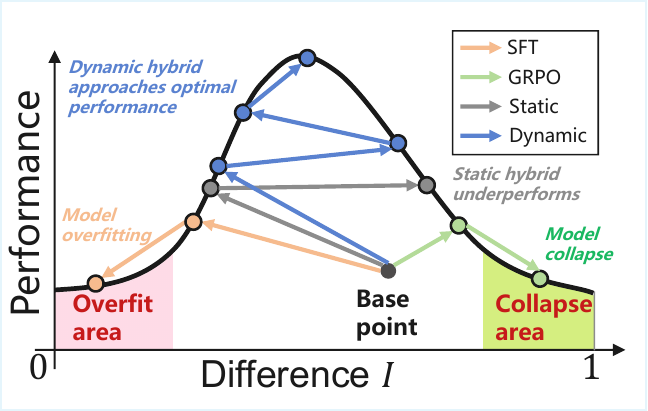}
    \caption{Advantages of dynamic training paradigm.\label{fig:dynamic:advantage}}
    \label{fig:theoritical}
    \end{subfigure}
    \caption{Visualizations of \sysname's architecture and theoretical analysis.}
\end{figure*}

In this section, we first present an overview of our proposed step-wise adaptive hybrid training framework, \sysname, which is inspired by the development of human reasoning abilities through structured learning followed by practice. We then theoretically analyze the advantages of \sysname over primitive training paradigms (i.e., SFT and GRPO) and static hybrid approaches, and further validate our insights through a series of case studies. Finally, we describe how \sysname adaptively adjusts the ratio between SFT and RL based on training dynamics after the warm-up phase (see Algorithm~\ref{algo:adaptive}).

\subsection{Overview of \sysname}
As illustrated in Figure~\ref{fig:arch}, \sysname consists of two components: a warm-up phase based on SFT and a subsequent hybrid training phase that integrates SFT and GRPO. We formally define these two phases below. These definitions serve as the foundation for the theoretical analysis in the next subsection.

\textbf{Warm-up phase.} Since \sysname targets task-specific scenarios in which high-quality datasets are often unavailable, it begins with SFT on a small-scale dataset of (\textit{question}, \textit{chain-of-thought}) pairs $(\boldsymbol{x}, \boldsymbol{e})$, where $\boldsymbol{x}$ denotes the input question token sequence and $\boldsymbol{e}$ denotes the corresponding chain-of-thought reasoning path that demonstrates the step-by-step solution process. The chain-of-thought is represented as a token sequence $\boldsymbol{e}=[a_1,\ldots,a_L\!=\!\texttt{<eos>}]$, where each $a_t$ corresponds to the $t$-th reasoning-step token generated autoregressively:

\begin{equation}
    a_t \sim \boldsymbol{\pi}_{\boldsymbol{\theta}}(\cdot|s_t), \quad s_{t+1} = [s_t, a_t],
\end{equation}

with $s_t$ denoting the state (context) at step $t$ and $\boldsymbol{\pi}_{\boldsymbol{\theta}}(\cdot|s_t)$ the token-generation probability conditioned on $s_t$.

During the SFT phase, the objective is to maximize the likelihood of the ground-truth sequences by minimizing the negative log-likelihood (NLL) loss:

\begin{equation}
    \mathcal{L}_{\text{SFT}}(\boldsymbol{\theta}) = -\mathbb{E}_{(\boldsymbol{x},\boldsymbol{e})\sim\mathcal{D}}\Bigl[\sum_{t=1}^{L}\log\boldsymbol{\pi}_{\boldsymbol{\theta}}(a_t|s_t)\Bigr],
\end{equation}

where $\mathcal{D}$ is the training data distribution, $\boldsymbol{\theta}$ the model parameters, and $L$ the target-sequence length.

\textbf{Hybrid training phase.}
After  the warm-up phase, \sysname performs  step-wise adaptive hybrid training employing both SFT and GRPO. In this phase, GRPO extends policy optimization through group-wise comparisons. For each input $q$, we sample $G$ outputs from the current and old policies. The objective combines advantage maximization with KL regularization to prevent excessive deviation:
\begin{align}
\mathcal{L}_{\text{GRPO}}(\theta)
&= \frac{1}{G}\sum_{i=1}^{G}
      \biggl[
        \min\!\left(
          \frac{\pi_\theta}{\pi_{\theta_{\text{old}}}}\hat{A}_{i,t},\;
          \text{clip}\!\left(
            \frac{\pi_\theta}{\pi_{\theta_{\text{old}}}},\,1\pm\epsilon
          \right)\hat{A}_{i,t}
        \right)
\nonumber\\
&\hspace{2em}
      - \beta D_{KL}\!\left[\pi_\theta \| \pi_{\text{ref}}\right]
      \biggr].
\label{eq:grpo_loss}
\end{align}
where $\pi_{\theta_{\text{old}}}$ is the previous policy before update, $\hat{A}_{i,t}$ denotes the estimated advantage value measuring how much better the action is compared to the average at that state,  $\pi_{\text{ref}}$ represents the reference policy (typically the initial SFT model), $\epsilon$ controls the clipping range for policy updates, and $\beta$ adjusts the strength of KL regularization. The ratio $\frac{\pi_\theta}{\pi_{\theta_{\text{old}}}}$ measures how much the new policy deviates from the old one for each action.

To formally define the dynamic adaptive training algorithm, we introduces $I(t)$ as the state function that returns the training paradigm decision variable $I(t)$ based on the current model's training state $t$. Unlike traditional hybrid methods that use a fixed training paradigm within an epoch, \sysname adopts a single training step $s$ as a training unit, enabling more flexible adaptive adjustments. Finally, we define the overall loss function $\mathcal{L}_{}(\theta)$ of the dynamic training swicth framework in Equation~\ref{eq:overall_loss}.

\begin{equation}
  \mathcal{L}_{}(\theta) = \frac{1}{S} \sum_{s=1}^{S} \left[ (1 - I(t)) \cdot \mathcal{L}_{\text{SFT}}(\theta) + I(t) \cdot \mathcal{L}_{\text{GRPO}}(\theta) \right]
\label{eq:overall_loss}
\end{equation}

\subsection{Theoretical Analysis of \sysname and Empirical Validation via Case Studies}

In this subsection, to theoretically examine the advantages of \sysname over existing training paradigms, we first establish the relationship between the gradient norm of the SFT loss and the Kullback–Leibler (KL) divergence. We then investigate how this relationship influences the reinforcement learning process. Specifically, we analyze the KL divergence between the model policy $\pi_\theta$ and the data distribution $\pi_{\text{data}}$, and how this divergence impacts the gradient norm of SFT.

Initially, we define the SFT loss function as the cross-entropy loss:
\begin{equation}
\mathcal{L}_{\text{SFT}} = -\mathbb{E}_{(x,y^*) \sim \mathcal{D}} \log \pi_\theta(y^* \mid x),
\end{equation}
where $\mathcal{D}$ represents the distribution of training data pairs $(x, y^*)$ consisting of input questions $x$ and their corresponding optimal reasoning paths $y^*$. The gradient of this loss with respect to model parameters $\theta$ is:
\begin{equation}
\nabla_\theta \mathcal{L}_{\text{SFT}} = -\mathbb{E}_{(x,y^*)} \nabla_\theta \log \pi_\theta(y^* \mid x).
\end{equation}
The KL divergence between the current policy $\pi_\theta$ and the data distribution $\pi_{\text{data}}$, which measures how much the model's behavior deviates from the original demonstrations, is given by:
\begin{equation}
D_{\text{KL}}(\pi_{\text{data}} \| \pi_\theta) = \mathbb{E}_{(x,y^*)} \log \frac{\pi_{\text{data}}(y^* \mid x)}{\pi_\theta(y^* \mid x)}.
\end{equation}
When $\pi_{\text{data}}(y^* \mid x)$ remains fixed during training (as is typical with human demonstrations), its gradient simplifies to:
\begin{equation}
\nabla_\theta D_{\text{KL}} = \mathbb{E}_{(x,y^*)} \nabla_\theta \log \pi_\theta(y^* \mid x) = \nabla_\theta \mathcal{L}_{\text{SFT}}.
\end{equation}
establishing the fundamental relationship shown in  Equation~\ref{eq:grad_kl_relation}:
\begin{equation}
\|\nabla_\theta \mathcal{L}_{\text{SFT}}\| =  \| \nabla_\theta D_{\text{KL}}(\pi_{\text{data}} \| \pi_\theta) \|.
\label{eq:grad_kl_relation}
\end{equation}

Therefore, we can derive:

\begin{equation}
D_{\text{KL}}(\theta')
= D_{\text{KL}}(\theta-\eta \nabla_\theta \mathcal{L}_{\text{SFT}})
\atop
\displaystyle \approx D_{\text{KL}}(\theta)
- \nabla_\theta D_{\text{KL}}(\theta)^{\!T}\eta\nabla_\theta \mathcal{L}_{\text{SFT}}
\displaystyle = D_{\text{KL}}(\theta)-\eta\|\nabla_\theta \mathcal{L}_{\text{SFT}}\|^2
\label{eq:kl}
\end{equation}

 This implies that by minimizing the SFT loss, \sysname effectively reduces the discrepancy between the model policy and the data distribution, thereby aligning the model policy more closely with the data distribution. As shown in Equation~\ref{eq:kl}, when the gradient norm of the SFT loss is large, a single training step effectively drives the model toward the data distribution. Conversely, when the gradient norm is small, the model risks overfitting or converging to a local optimum. In such cases, incorporating the reinforcement signal from GRPO helps escape local optima by encouraging exploration of high-reward regions, thereby balancing stability and generalization.
 
In GRPO, KL divergence constrains the policy to remain close to a reference model, thereby mitigating mode collapse. In small-scale models, removing the KL penalty eliminates the additional training overhead caused by gradient conflict and capacity occupation, enabling more efficient distribution shift and yielding stronger task-specific performance.  

Nevertheless, unconstrained reinforcement learning is prone to catastrophic forgetting during policy updates. We therefore integrate KL-divergence–based SFT supervision with GRPO and, by continuously monitoring training dynamics, dynamically balance free exploration against stable constraints while fully leveraging the training data. Below, we explain why \sysname outperforms existing paradigms, as illustrated in Figure~\ref{fig:dynamic:advantage}.

\textbf{Avoiding SFT-induced overfitting.}
Recent studies demonstrate that pure SFT overfits to limited CoT   demonstrations~\cite{luong2024reftreasoningreinforcedfinetuning}. GRPO’s exploration of diverse reasoning paths ($G$ samples per prompt) breaks this limitation.      

\textbf{Mitigating model collapse caused by RL.} Standard RL exhibits mode collapse and reward hacking. When the LLM deviates substantially from the original data distribution, our theory shows that SFT pulls the policy back toward the required training distribution. The hybrid approach maintains proximity to this distribution by injecting SFT constraints whenever deviation is detected, preventing degenerate solutions while still permitting reward-guided exploration beyond pure SFT. The non-negativity of KL divergence guarantees that the policy never drifts too far from the data distribution, thus averting mode collapse during policy updates.

\textbf{Overcoming the suboptimality of static hybrid training.}
Our gradient-based adaptation:

\begin{equation}
p_t = \left( \frac{\|\nabla D_{\text{KL}}(\theta)^t\| }{\|\nabla D_{\text{KL}}(\theta)^t\|+ \gamma\|\nabla D_{\text{KL}}(\theta)^0\|}\right)
\end{equation}

where $\|\nabla D_{\text{KL}}(\theta)^t\|$ denotes the KL divergence between the current training step's distribution and the original data distribution, and $\|\nabla D_{\text{KL}}(\theta)^0\|$ denotes  the  KL divergence at the final training step of the warm-up phase. $\sigma$ is the sigmoid function, provides smooth transitions between:
\begin{itemize}
    \item \textit{Exploration-dominant} phase ($p_t \to 1$):  $\|\nabla D_{\text{KL}}(\theta)^t\|$ is relatively large, meaning that LLM is currently far from the original distribution of the data and requires enhanced supervised learning.
    \item \textit{Exploitation-dominant} phase ($p_t \to 0$): $\|\nabla D_{\text{KL}}(\theta)^0\|$ is relatively large, meaning that LLM is currently close to the original distribution of the data and requires enhanced exploration.
\end{itemize}

\begin{figure}[h]
\centering
\includegraphics[width=\linewidth]{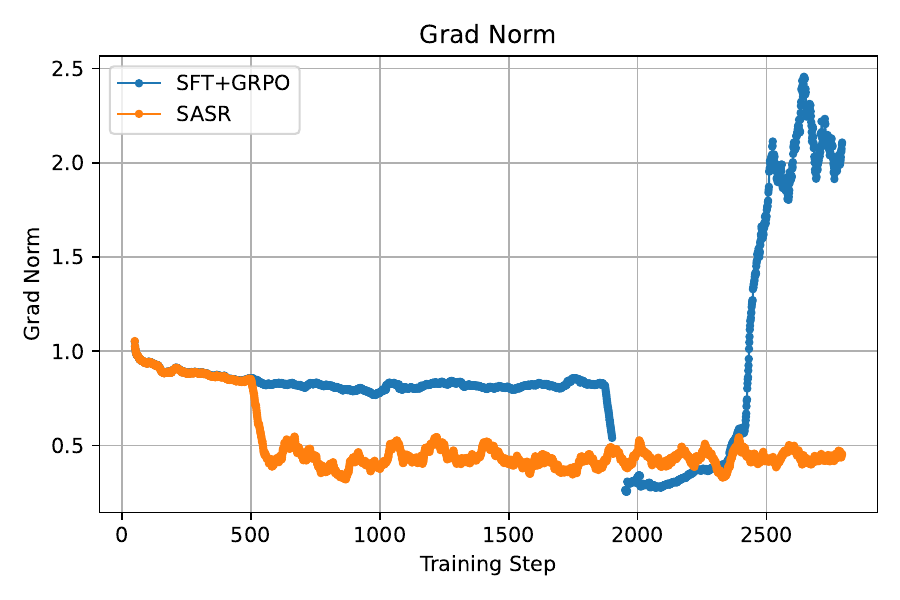}
\caption{Gradient norm under different paradigms.}
\label{fig:case1}

\end{figure}

\begin{figure}[h]
\centering
\includegraphics[width=\linewidth]{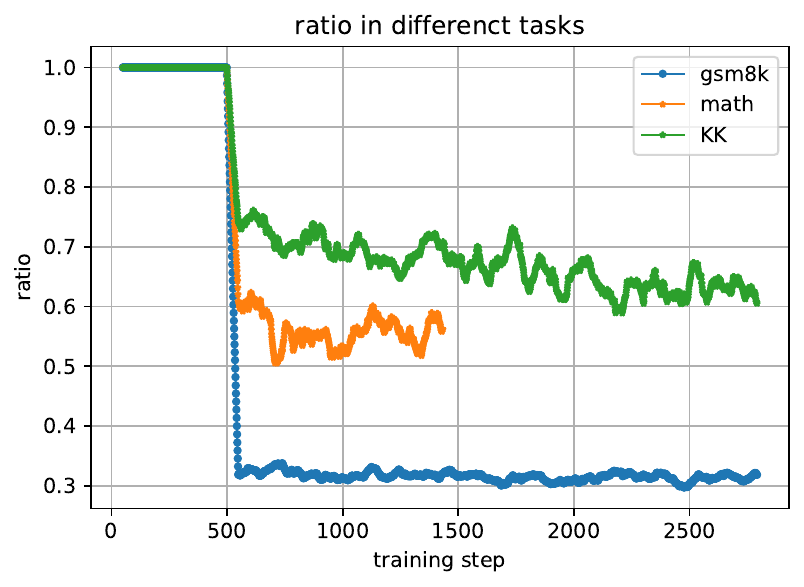}
\caption{Visualizations of cases study in theoretical analysis of \sysname.}
\label{fig:case2}

\end{figure}

Relative to static-switching hybrids, \sysname offers three advantages. First, inspired by human curriculum learning, it decomposes mixed training into multiple steps tuned to training state and performs SFT and exploration simultaneously, resolving the \textit{catastrophic forget–stagnation trade-off}. Second, it smooths the transition between regimes, reducing the negative impact of abrupt switches (Figure~\ref{fig:case1}). Third, as Figure~\ref{fig:case2} reveals, the optimal warm-up duration and switching point vary across datasets; static schedules therefore generalize poorly, whereas \sysname adapts automatically.

\subsection{Dynamic Ratio Selection}

\begin{algorithm}[t]
\caption{SASR: Step-wise Adaptive SFT-RL Training}
\begin{algorithmic}[1]
\REQUIRE $\mathcal{D}_{\text{train}} = \{(\boldsymbol{x}, \boldsymbol{e}, \boldsymbol{y})\}$: Tuples of (question, CoT, answer), $W$: warm-up steps, $T$: total steps, $\boldsymbol{\pi}_\theta^{(0)}$: initial policy, $G$: group size
\ENSURE $\boldsymbol{\pi}_{\boldsymbol{\theta}}$: final policy

\STATE Initialize $\boldsymbol{\pi}_{\boldsymbol{\theta}} = \boldsymbol{\pi}_{\boldsymbol{\theta}}^{(0)}$
\STATE // Warm-up stage
\FOR{$i = 1$ to $W$}
    \STATE Sample $(\boldsymbol{x}, \boldsymbol{e}, \boldsymbol{y}) \sim \mathcal{D}_{\text{train}}$
    \STATE Update $\boldsymbol{\theta} = \textsc{Optimization\_Step}(\mathcal{L}_{\text{SFT}}(\boldsymbol{\theta}))$
    \IF{$i = W$}
        \STATE $G_{\text{warmup}} \gets \|\nabla_{\boldsymbol{\theta}} \mathcal{L}_{\text{SFT}}(\boldsymbol{\theta})\|$
    \ENDIF
\ENDFOR

\STATE // Adaptive training stage
\FOR{$t = 1$ to $T$}
    \STATE Compute $p = \frac{G_{\text{last-SFT}}}{G_{\text{last-SFT}} + \gamma G_{\text{warmup}}}$
    \STATE Sample $\alpha \sim \text{Uniform}(0,1)$
    \IF{$\alpha < p$}
        \STATE Sample $(\boldsymbol{x}, \boldsymbol{e}, \boldsymbol{y}) \sim \mathcal{D}_{\text{train}}$
        \STATE Update $\boldsymbol{\theta} = \textsc{Optimization\_Step}(\mathcal{L}_{\text{SFT}}(\boldsymbol{\theta}))$
        \STATE $G_{\text{last-SFT}} \gets \|\nabla_{\boldsymbol{\theta}} \mathcal{L}_{\text{SFT}}(\boldsymbol{\theta})\|$
    \ELSE
        \STATE Sample $(\boldsymbol{x}, \_, \boldsymbol{y}) \sim \mathcal{D}_{\text{train}}$
        \STATE Generate $\{\hat{\boldsymbol{e}}_i\}_{i=1}^G \sim \boldsymbol{\pi}_{\boldsymbol{\theta}}(\boldsymbol{x})$
        \STATE Extract $\{\hat{\boldsymbol{y}}_i\}_{i=1}^G \gets \textsc{Extract}(\{\hat{\boldsymbol{e}}_i\})$
        \STATE Compute rewards $\{R(\hat{\boldsymbol{y}}_i)\}_{i=1}^G$
        \STATE Form groups $\mathcal{G}_+, \mathcal{G}_-$ based on reward percentiles
        \STATE Update $\boldsymbol{\theta} = \textsc{Optimization\_Step}(\mathcal{L}_{\text{GRPO}}(\boldsymbol{\theta}))$
    \ENDIF
\ENDFOR

\RETURN $\boldsymbol{\pi}_{\boldsymbol{\theta}}$
\end{algorithmic}
\label{algo:adaptive}
\end{algorithm}

Our training framework integrates SFT with GRPO through an adaptive mechanism whose core innovation lies in dynamically balancing the two paradigms according to the model’s live performance metrics.  
Designing a reliable switching indicator for this SFT-RL hybrid remains the central challenge. Beyond naive rule-based schedules, we systematically examine training-state signals such as gradient norm and step count. 

\textbf{SSR: rule-based training schedule.} We first introduce a naive baseline—Step-wise SFT–RL (SSR)—that alternates SFT and GRPO at every training step, granting each method equal step counts. While ensuring balanced exposure, SSR lacks any mechanism to modulate the relative contribution of SFT and GRPO, inviting either overfitting or collapse.

\textbf{SSR\_cosine: cosine training Schedule}. SSR\_cosine refines this approach by scheduling the probability of selecting an SFT step via a cosine decay. Aligned with the curriculum-learning-quiz perspective, SFT dominates early training until the model masters response formats and logical patterns; thereafter the schedule gradually favors GRPO. The limitation is its blindness to actual training states, rendering it unable to react to dynamic deviations.

\begin{equation}
I(t) = 0.5 \left(1 + \cos\left(\pi \frac{s}{S}\right)\right) (U - L) + L,
\end{equation}
where s represents the numbers of current step, S represents the maximum number in training process, and U and L respectively represent the predefined upper and lower bounds parameters.

\textbf{The adaptive training schedule ultimately adopted by SASR.}
To ensure cross-task generalization, \sysname employs an adaptive algorithm (Algorithm~\ref{algo:adaptive}) that derives its switching signal from the gradient norm of the KL divergence between the current policy $\pi_\theta$ and the original data distribution $\pi_{\text{data}}$. At each step the algorithm compares the running KL divergence with the warm-up benchmark norm and computes a mixing weight that decides between SFT and GRPO.  
When $D_{\text{KL}}(\pi_\theta\|\pi_{\text{data}})$ is large relative to the benchmark, \sysname raises the SFT weight, safeguarding foundational reasoning. When the divergence is small, \sysname increases the GRPO weight, shifting the regime toward “quiz” mode: the policy samples $G$ diverse trajectories, computes relative advantages via grouped comparisons, and updates with the truncated GRPO objective (Equation~\ref{eq:grpo_loss}).

\section{Experimental Results}

\subsection{Experimental settings}

\textbf{Dataset.} We gather three representative datasets for the experiment: GSM8K~\cite{cobbe2021gsm8k}, KK~\cite{xie2025logicrlunleashingllmreasoning}, and MATH~\cite{hendrycksmath2021}. The GSM8K dataset comprises elementary-level mathematical problems that mainly demand arithmetic computation. A distinctive characteristic of this dataset is that the final answers are confined to non-negative integers, and the format requirements for response presentation are relatively lenient. MATH contains more challenging competition-style problems whose answers include mathematical formulas. In contrast, KK functions as a logic-reasoning benchmark designed specifically to assess models' deductive abilities. Unlike GSM8K and MATH, KK enforces highly specific output-format requirements to guarantee unambiguous and complete responses.

\begin{table}[!htbp]
    
    \centering
    \small
    \setlength{\tabcolsep}{3pt}
    \begin{tabular}{ll|lll|l}
        \toprule
        Model& $|\boldsymbol{\theta}|$   & GSM8K & MATH & KK & Avg.\\
        \midrule
        GPT-4o& 200B   & 0.818 &0.620&0.33& 0.589\\ 
        Deepseek-V3& 671B & 0.908 &0.870&0.57& 0.783 \\
        \midrule
        Baseline& 1.5B/~0.5B & 0.638 & 0.146&0.03& 0.271 \\
        SFT & 1.5B/~0.5B & 0.752 & 0.212&0.28& 0.414 \\
        GRPO & 1.5B/~0.5B & 0.557 & 0.170 &0.09& 0.272\\
        Static hybrid & 1.5B/~0.5B & \textbf{0.814} & 0.160 &0.33& 0.435 \\
        SSR & 1.5B/~0.5B & 0.779 & 0.196 &0.38& 0.452 \\
        SSR\_cosine& 1.5B/~0.5B  & 0.795 & 0.204 &0.39& 0.463\\
        SASR & 1.5B/~0.5B & 0.803 & \textbf{0.230} &\textbf{0.42} &\textbf{0.484} \\
        \bottomrule
    \end{tabular}
    \caption{Answer accuracy of different models on the task-specific problems.}
    \label{tab:math}
\end{table}

\begin{table*}[t]
  
  \centering
  
  \renewcommand{\arraystretch}{0.9}
  {
  \begin{tabular}{l|lllllll|l}
    \toprule
    \  & \multicolumn{8}{c}{difficulty level}   \\
    \cmidrule(r){2-9} 
    Model     & 2 ppl  & 3 ppl & 4 ppl & 5 ppl & 6 ppl & 7 ppl & 8 ppl  & Avg. \\
    \midrule
    Deepseek-Math-7B-Instruct & 0.12 & 0.08 & 0.05 & 0.02 & 0.00 & 0.00 & 0.00 & 0.04  \\
    NuminaMath-7B-CoT  & 0.14 & 0.03 & 0.02 & 0.00 & 0.00 & 0.00 & 0.00 & 0.03  \\
    Qwen2.5-Base-7B  & 0.23 & 0.11 & 0.06 & 0.05 & 0.02 & 0.01 & 0.00 & 0.07  \\
    Qwen2.5-7B-Instruct-1M & 0.23 & 0.25 & 0.13 & 0.09 & 0.03 & 0.04 & 0.02 & 0.11  \\
    GPT-4o & 0.73 & 0.46 & 0.40 & 0.31 & 0.19 & 0.12 & 0.07 & 0.33\\
    Deepseek-V3 & 0.90 & 0.71 & 0.68 & 0.57 & 0.39 & 0.37 & 0.37 & 0.57\\ 
    \midrule
    Qwen2.5-1.5B-Instruct & 0.13 & 0.05 & 0.02 & 0.00 & 0.00 & 0.00 & 0.00 & 0.03 \\
    SFT & 0.65 & 0.42 & 0.38 & 0.24 & 0.14 & 0.08 & 0.07 & 0.28 \\
    GRPO & 0.32 & 0.17 & 0.07 & 0.05 & 0.03 & 0.01 & 0.01 & 0.09 \\
    Static hybrid & 0.71 & 0.47 & 0.37 & 0.25 & \textbf{0.26} & 0.10 & 0.12 & 0.33 \\
    SSR & \textbf{0.84} & 0.54 & 0.47 & 0.34 & 0.19 & \textbf{0.15} & 0.12 & 0.38 \\
    SSR\_cosine & 0.74 & 0.60 & 0.49 & 0.36 & 0.22 & 0.14 & 0.15 & 0.39 \\
    SASR & 0.83 & \textbf{0.68} & \textbf{0.55} & \textbf{0.39} & 0.19 & 0.11 & \textbf{0.17} & \textbf{0.42} \\
    \bottomrule
  \end{tabular}}
  \caption{Answer accuracy of different models on the Knight-and-Knives problem with various level of difficulty. }
  \label{k&k experiment}
\end{table*}

\textbf{Model Selection.}  
We select the 1.5B and 0.5B variants from the widely-adopted Qwen2.5-Instruct series as our backbone models to demonstrate the effectiveness and generalization of \sysname on small-scale training tasks. According to the official technical report~\cite{qwen}, both models exhibit solid capabilities in chain-of-thought reasoning and instruction following. Our evaluation protocol directly uses the widely-validated open-source QwenLM implementation, ensuring full alignment with current mainstream practices~\cite{qwen,xie2025logicrlunleashingllmreasoning}.

\textbf{Model Training}: 
We employ model distillation to normalize the dataset format and enhance the quality of the CoT. As Wadhwa et al.~\cite{wadhwa2024investigating} found out, the quality of CoT could also affect SFT results. To address these limitations, we apply rejective sampling method to aquire high quality of the CoT in dataset. We  distill large language models to generete CoT instead of standard solutions. For GSM8K, we obtain high-quality CoT annotations from the gsm8k\_distilled dataset provided by Camel-AI.  For MATH, we distill CoT from the Qwen2.5-Math-1.5B model and filter them according to the correctness of the answers. Consistent performance improvements are observed when we replace standard solutions with distilled CoT. Referring to the static hybrid paradigm adopted by the advanced LLM DeepSeek-R1~\cite{deepseekai2025deepseekr1incentivizingreasoningcapability}, in the static hybrid training, we switch training methods (SFT \& RL) on a per-epoch basis, specifically conducting 2 epochs of SFT followed by 1 epoch of GRPO. For the sake of reproducibility, we provide additional training and evaluation details in the appendix.

\textbf{Details of resources used}
 Every model is trained on two NVIDIA H20 141GB GPUs. We use automatic mixed precision (BFloat16) for two math datasets and single precision (Float32) for the KK dataset. In total, it takes around 45 GPU days to complete all experiments listed in Table 1 on NVIDIA H20 141GB.

\subsection{Experimental Results}

\subsubsection{Performance on Mathematical Reasoning Tasks}

We trained models on Mathematical Reasoning Tasks respectively to evaluate our method, and we follow the evaluation methods of reasoning tasks from QwenLM~\cite{qwen}. The main results are shown in Table~\ref{tab:math}. In the commonly used benchmark tests of mathematical reasoning, the classical training paradigm SFT can enhance the ability of the model, but the improvement is limited. However, the sole use of RL (GRPO) has caused the degradation of the ability of the base model due to the problem of pattern collapse. Hybrid training can further enhance the reasoning ability of the model. Among them, the performance of the designed training schedule is superior to that of direct hybrid (SSR). It is observed that \sysname significantly enhanced the DeepSeek-R1-Distill-Qwen-1.5B model, increasing the accuracy rate from 63.8\% to 80.3\%, reaching a level close to GPT-4o. Due to carefully designed CoT distillation, SFT achieved remarkably improvement on the MATH dataset. The experimental results provide empirical evidence that \sysname further exceeds SFT, with a measurable improvement of 1.8\%.


\subsubsection{Performance on Logical Inference Tasks}

For the KK dataset, the models are trained on 3 to 7-person KK problems and evaluated on 2 to 8-person KK problems. We exclude 2 person and 8 person problems from training datasets to observe whether the model could generalize to those two cases. We follow the base evaluation method of KK dataset~\cite{xie2025logicrlunleashingllmreasoning} to decide whether the response of the model is accurate. The results of our methods and other baseline models are in Table~\ref{k&k experiment}. Our experiments suggest that  \sysname  achieved better results compared to SFT, GRPO and static hybrid training paradims. Consequently, \sysname has an average accuracy improvement of 9\% compared to GPT-4o. This demonstrates that \sysname yields superior training performance on logical inference tasks compared to pure SFT, RL, and existing static hybrid methods.

\subsubsection{Ablation Experiment}
To validate the contribution of each design component in \sysname, we conduct ablation studies on GSM8K, MATH, and KK. Pure SFT underperforms on challenging data (75.2\%, 21.2\%, 28\%), revealing overfitting; GRPO begins evenly but degrades due to reward hacking and mode collapse. Models trained with Pure SFT or GRPO consistently underperform compared to the various hybrid scheduling schemes, demonstrating that any single paradigm alone is insufficient to simultaneously ensure stability and generalization, whereas hybrid scheduling effectively harnesses their complementary strengths.
static hybrids suffer cross-task instability from rigid switching, achieving only 16.0\% on MATH. SSR and SSR\_cosine temper abrupt transitions yet ignore live training state, yielding 45.2\% and 46.3\% versus \sysname’s 48.4\%, underscoring the superiority of dynamic adaptation.

\subsection{Time cost analysis}
\begin{figure}[h]
\centering
\includegraphics[width=\linewidth]{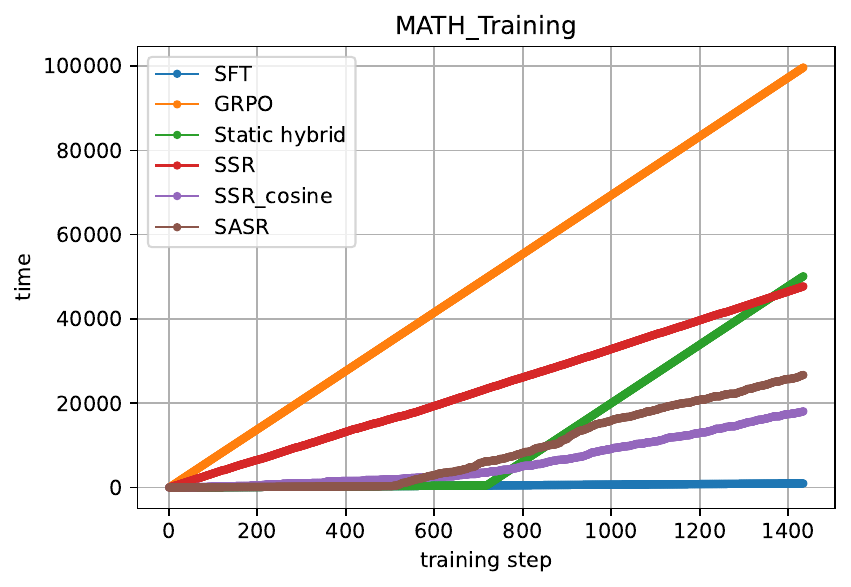}
\caption{Visualizations of time cost in MATH dataset}
\label{fig:ts}

\end{figure}

Considering that our \sysname may have cost issues as a training paradigm, we conducted a Time Cost Analysis of \sysname. We performed a complete training process using different training paradigms on the MATH dataset and recorded the time for each training step, which is visualized in Figure~\ref{fig:ts}. Our analysis shows that the time required for direct GRPO is significantly longer than that for direct SFT. This difference is due to the distinct processes of computing the loss functions in the two methods. Specifically, SSR\_cosine outperforms our \sysname in terms of time cost. However, the training process of SSR\_cosine is related to the hyperparameter Settings (RL ratio), and it is unstable in both training effect and time cost. Compared with the simple combination of GRPO and SFT (SSR, Static hybrid), our method (\sysname) significantly reduces the training time while achieving better performance. This indicates that our method is superior to traditional training paradigms in terms of both time cost and performance improvement.

\section{Conclusion and Future Work}
In this work, we theoretically connect SFT and GRPO by modeling human reasoning cultivation, proposing a step-wise adaptive hybrid training framework for task-specific LLMs. \sysname outperforms SFT, RL, and static hybrid methods on GSM8K, MATH, and KK datasets in reasoning tasks. By monitoring training status and step-level adjustment, \sysname ensures smooth transitions between schemes while maintaining core reasoning abilities. The framework further reduces training time relative to naive hybrids, and its gradient-based switching mechanism generalises across domains.

However, \sysname still has certain limitations. First, its compatibility with other reinforcement-learning algorithms (e.g., PPO, DAPO) within the hybrid framework remains to be validated. Second, its applicability to broader LLM domains, such as question answering, requires further investigation. Finally, the current use of task-specific hyper-parameter settings across different dataset categories may hinder the generalizability of \sysname; we will explore optimal hyper-parameter configurations in future work.

\bibliography{aaai2026}

\clearpage

\appendix


\section{Experimental Details}
\subsection{Model and Training Hyperparameters}
The appendix provides a complete compilation of the training hyperparameters for all compared models on the GSM8K, MATH, and KK datasets.
Tables~\ref{hyp:gsm8k}, \ref{hyp:math}, and \ref{hyp:KK} list the key hyperparameters for each model on GSM8K, MATH, and KK, respectively; any secondary hyperparameters omitted from the tables are kept identical to their official implementations.

\label{app:hyper}

\begin{table}[htb]\centering

\raggedright{\textbf{Parameter Setting (GSM8K)}}
\resizebox{\linewidth}{!}{
\begin{tabular}{p{0.3\linewidth}p{0.3\linewidth}p{0.3\linewidth}}
\midrule[0.3ex]
\textbf{Method} &
\textbf{Parameters} \\
\midrule[0.15ex]
Baseline
& 
base model: DeepSeek-R1-Distill-Qwen-1.5B 
&
learning rate: 2e-5 \newline
\\
\midrule[0.15ex]
SFT
& 
batch size: 16 \newline
epoch: 3 
&
learning rate: 2e-5 \newline
 \\
\midrule[0.15ex]
GRPO
& 
batch size: 16 \newline
beta: 0.04 
&
learning rate: 2e-5 \newline
epoch: 3 
\\
\midrule[0.15ex]
SSR
& 
batch size: 16 \newline
beta: 0.04 
&
learning rate: 2e-5 \newline
epoch: 3 
\\
\midrule[0.15ex]
SSR\_cosine
& 
batch size: 16 \newline
beta: 0.04 \newline
lower\_limit: 0.1
&
learning rate: 2e-5 \newline
epoch: 3 \newline
upper\_limit: 0.9 \\
\midrule[0.15ex]
SASR
& 
batch size: 16 \newline
beta: 0.04 \newline
warmup\_steps: 500
&
learning rate: 2e-5 \newline
epoch: 3 \newline
\\
\midrule[0.3ex]

\end{tabular}}
\caption{Parameter Setting (GSM8K)}
\label{hyp:gsm8k}
\end{table}

\begin{table}[htb]\centering

\raggedright{\textbf{Parameter Setting (MATH)}}
\resizebox{\linewidth}{!}{
\begin{tabular}{p{0.3\linewidth}p{0.3\linewidth}p{0.3\linewidth}}
\midrule[0.3ex]
\textbf{Method} &
\textbf{Parameters} \\
\midrule[0.15ex]
Baseline
& 
base model: Qwen2.5-0.5b-Instruct 
&
learning rate: 1e-5 \newline
 \\
\midrule[0.15ex]
SFT
& 
batch size: 16  \newline
epoch: 2
&
learning rate: 1e-5 
 \\
\midrule[0.15ex]
GRPO
& 
batch size: 16 \newline
beta: 0.04 
&
learning rate: 1e-5 \newline
epoch: 2 
 \\
\midrule[0.15ex]
SSR
& 
batch size: 16 \newline
beta: 0.04 
&
learning rate: 1e-5 \newline
epoch: 2 
 \\
\midrule[0.15ex]
SSR\_cosine
& 
batch size: 16 \newline
beta: 0.04 \newline
lower\_limit: 0.6
&
learning rate: 1e-5 \newline
epoch: 2 \newline
upper\_limit: 0.99 \\
\midrule[0.15ex]
SASR
& 
batch size: 16 \newline
beta: 0.04 \newline
warmup\_steps: 500
&
learning rate: 1e-5 \newline
epoch: 2
 \\
\midrule[0.3ex]

\end{tabular}}
\caption{Parameter Setting (MATH)}
\label{hyp:math}
\end{table}

\begin{table}[htb]\centering

\raggedright{\textbf{Parameter Setting (KK)}}
\resizebox{\linewidth}{!}{
\begin{tabular}{p{0.3\linewidth}p{0.3\linewidth}p{0.3\linewidth}}
\midrule[0.3ex]
\textbf{Method} &
\textbf{Parameters} \\
\midrule[0.15ex]
Baseline
& 
base model: Qwen2.5-1.5b-Instruct 
&
learning rate: 1e-5 \newline
 \\
\midrule[0.15ex]
SFT
& 
batch size: 16 \newline
epoch: 3 
&
learning rate: 1e-5 
 \\
\midrule[0.15ex]
GRPO
& 
batch size: 16 \newline
beta: 0.04 
&
learning rate: 1e-5 \newline
epoch: 3 
 \\
\midrule[0.15ex]
SSR
& 
batch size: 16 \newline
beta: 0.04 
&
learning rate: 1e-5 \newline
epoch: 3 
 \\
\midrule[0.15ex]
SSR\_cosine
& 
batch size: 16 \newline
beta: 0.04 \newline
lower\_limit: 0.6
&
learning rate: 1e-5 \newline
epoch: 3 \newline
upper\_limit: 0.99 \\
\midrule[0.15ex]
SASR
& 
batch size: 16 \newline
beta: 0.04 \newline
warmup\_steps: 500
&
learning rate: 1e-5 \newline
epoch: 3 
 \\
\midrule[0.3ex]

\end{tabular}}
\caption{Parameter Setting (KK)}
\label{hyp:KK}
\end{table}

\subsection{Optimization and Schedule}
We trained the model using the AdamW optimizer with the following hyperparameters: weight decay $\lambda = 0.0$, $\beta_1 = 0.9$, $\beta_2 = 0.999$, $\epsilon = 10^{-8}$, and batch size $B = batch size$ per device with gradient accumulation steps . 
The training ran for  epochs with a maximum gradient norm of $1.0$. 
We used a linear learning rate scheduler.

\subsection{Reward Setting}
Below, we present the prompts used for GSM8K, MATH, KK. 

\textbf{GSM8K}:

The \texttt{correctness\_reward}  checks the correctness of the answer and assigns rewards based on the match between the extracted answer and the expected answer.

\begin{equation}
R_{\text{correctness}} = 
\begin{cases} 
2.0 & \text{if } \text{float}(\text{extracted}) = \text{float}(\text{answer}) \\
0.0 & \text{otherwise}
\end{cases}
\end{equation}

The \texttt{int\_reward}  checks whether the answer is an integer and assigns rewards based on whether the answer is an integer.

\begin{equation}
R_{\text{integer}} = 
\begin{cases} 
0.5 & \text{if } \text{answer.isdigit()} \\
0.0 & \text{otherwise}
\end{cases}
\end{equation}

The \texttt{combined\_format\_reward}  checks the format of the answer and assigns rewards based on the format of the answer.

\begin{equation}
R_{\text{format}} = 
\begin{cases} 
1.0 & \text{if perfect\_match} \\
0.5 & \text{if answer\_matches} \\
0.0 & \text{otherwise}
\end{cases}
\end{equation}

\begin{equation}
\text{if remaining\_text:}
\begin{cases} 
\text{penalty} = \min(0.4, \text{len}(\text{remaining\_text}) \times 0.01) \\
R_{\text{format}} = \max(R_{\text{format}} - \text{penalty}, 0.1)
\end{cases}
\end{equation}
The final reward is the sum of the above-mentioned rewards.

\textbf{MATH}:

The \texttt{correctness\_reward} checks the correctness of the answer and assigns rewards based on the match between the extracted answer and the expected answer.

\begin{equation}
R_{\text{correctness}} = 
\begin{cases} 
2.0 & \text{if } \text{is\_equivalent}(\text{extracted}, \text{answer}) \\
0.0 & \text{otherwise}
\end{cases}
\end{equation}

The \texttt{format\_reward} checks the format of the answer and assigns rewards based on the format of the answer.

\begin{equation}
R_{\text{format}} = 
\begin{cases} 
1.0 & \text{if } \text{perfect\_match} \\
0.0 & \text{otherwise}
\end{cases}
\end{equation}

The final reward is the sum of the above-mentioned rewards.

\textbf{KK}:

The \texttt{compute\_score} function evaluates model responses through two main reward components:

The \texttt{format\_reward} checks the XML-style tag structure and assigns rewards based on proper formatting:

\begin{equation}
R_{\text{format}} = 
\begin{cases} 
1.0 & \text{if } \text{validate\_response\_structure()} = \text{True} \\
-1.0 & \text{otherwise}
\end{cases}
\end{equation}

where validation requires:
\begin{itemize}
\item Exactly one pair of \texttt{\texttt{<think>}} and \texttt{\texttt{</think>}} tags
\item Exactly one pair of \texttt{\texttt{<answer>}} and \texttt{\texttt{</answer>}} tags
\item Correct nesting order: \texttt{\texttt{<think>}} before \texttt{\texttt{<answer>}}
\end{itemize}

The \texttt{answer\_reward} evaluates solution accuracy with tiered scoring:

\begin{equation}
R_{\text{correctness}} = 
\begin{cases} 
2.0 & \text{if } \text{pred\_status} = \text{gt\_status} \\
-1.5 & \text{if complete but incorrect answer} \\
-2.0 & \text{if incomplete answer or format failure}
\end{cases}
\end{equation}

Answer parsing requires:
\begin{itemize}
\item Number of knight/knave declarations matches character count
\item Explicit role assignments for all expected names
\item Proper \texttt{b(knight)b} or \texttt{b(knave)b} notation
\end{itemize}

The final reward is the sum of the format and answer rewards.

\section{Simulator Details}
The appendix compiles the inference prompts used by all models on the GSM8K, MATH, and KK datasets.  
Tables~\ref{tab:gsm8k_prompt}, \ref{tab:math_prompt}, and \ref{tab:kk_prompt} present the specific prompt templates for GSM8K, MATH, and KK, respectively.

\begin{table}[htb]
\centering
\raggedright{\textbf{Prompt for GSM8K Dataset}}

\resizebox{\linewidth}{!}{%
\begin{tabular}{p{1.1\linewidth}}
\toprule[0.3ex]
\#\#\# System\newline
You are a helpful assistant that solves arithmetic word problems. For each problem, you should first analyze the problem step by step, perform the necessary calculations, and then provide the final answer. Present your reasoning process before giving the final answer, which should be prefixed with "The answer is".\newline

\#\#\# Demonstration Examples\newline
\textbf{Example 1:}\newline
Question: There are 15 trees in the grove. Grove workers will plant trees in the grove today. After they are done, there will be 21 trees. How many trees did the grove workers plant today?\newline
Let's think step by step\newline
There are 15 trees originally. Then there were 21 trees after some more were planted. So there must have been 21 - 15 = 6.\newline
The answer is 6\newline

\textbf{Example 2:}\newline
Question: If there are 3 cars in the parking lot and 2 more cars arrive, how many cars are in the parking lot?\newline
Let's think step by step\newline
There are originally 3 cars. 2 more cars arrive. 3 + 2 = 5.\newline
The answer is 5\newline

\#\#\# User\newline
\{\{problem\_statement\}\}\newline

\#\#\# Assistant\newline
Let's think step by step\newline
\{\{step-by-step reasoning\}\}\newline
The answer is \{\{final numerical answer\}\}
\\ \bottomrule[0.3ex]
\end{tabular}%
}
\caption{Prompt for GSM8K Dataset}
\label{tab:gsm8k_prompt}
\end{table}

\begin{table}[htb]\centering
\raggedright{\textbf{Prompt for MATH Dataset}}

\resizebox{\linewidth}{!}{
\begin{tabular}{p{1.1\linewidth}} 
\toprule[0.3ex]

\#\#\# System\newline
You are a helpful assistant that solves grid-based path counting problems. For each problem, you should first analyze the movement constraints and possible paths systematically, then provide the final count of distinct paths. Present your reasoning process within \texttt{<think>} tags and the final numerical answer within \texttt{<answer>} tags.\newline

\#\#\# Example User Question\newline
On the x,y-plane, the origin is labeled with an N. The points $(1,0)$, $(-1,0)$, $(0,1)$, and $(0,-1)$ are labeled with A's. The points $(2,0)$, $(1,1)$, $(0,2)$, $(-1,1)$, $(-2,0)$, $(-1,-1)$, $(0,-2)$, and $(1,-1)$ are labeled with T's. The points $(3,0)$, $(2,1)$, $(1,2)$, $(0,3)$, $(-1,2)$, $(-2,1)$, $(-3,0)$, $(-2,-1)$, $(-1,-2)$, $(0,-3)$, $(1,-2)$, and $(2,-1)$ are labeled with H's. If you are only allowed to move up, down, left, and right, starting from the origin, how many distinct paths can be followed to spell the word NATH?\newline

\#\#\# Example Assistant Response\newline
\texttt{<think>}
1. From N at (0,0), we can move to 4 adjacent A's at (±1,0) and (0,±1)\newline
2. From any A, we can move to 3 T's (considering movement constraints)\newline
3. From side T's, we can reach 2 H's each; from bottom T, we can reach 3 H's\newline
4. Total paths per A: 2×2 (side T's) + 3 (bottom T) = 7 paths\newline
5. Multiply by 4 A's: 4 × 7 = 28 total paths
\texttt{</think>}\newline
\texttt{<answer>}28\texttt{</answer>}

\#\#\# User\newline
\{\{problem\_statement\}\}\newline

\#\#\# Assistant\newline
\texttt{<think>}
\{\{step-by-step reasoning\}\}
\texttt{</think>}\newline
\texttt{<answer>}
\{\{final numerical answer\}\}
\texttt{</answer>}

\\ \bottomrule[0.3ex]
\end{tabular}
}
\caption{Prompt for MATH Dataset}
\label{tab:math_prompt}
\end{table}

\begin{table}[htb]\centering
\raggedright{\textbf{Prompt for KK dataset}}

\resizebox{\linewidth}{!}{
\begin{tabular}{p{1.1\linewidth}} 
\toprule[0.3ex]

\#\#\# System\newline
You are a helpful assistant. The assistant first thinks about the reasoning process in the mind and then provide the user with the answer. The reasoning process and answer are enclosed within \texttt{<think>} \texttt{</think>} and \texttt{<answer>} \texttt{</answer>} tags, respectively, i.e., \texttt{<think>}reasoning process here\texttt{</think>}\texttt{<answer>}answer here\texttt{</answer>}. Now the user asks you to solve a logical reasoning problem. After thinking, when you finally reach a conclusion, clearly state the identity of each character within \texttt{<answer>} tags, i.e., \texttt{<answer>} (1) ...\newline(2) ...\texttt{</answer>}.\newline

\#\#\# User\newline
\{\{problem\_statement\}\}\newline

\#\#\# Assistant\newline
\texttt{<think>}
\{\{reasoning\_steps\}\}
\texttt{</think>} \newline
\texttt{<answer>}
\{\{final\_answer\}\}
\texttt{</answer>}

\\ \bottomrule[0.3ex]
\end{tabular}
}
\caption{Prompt for KK Dataset}
\label{tab:kk_prompt}
\end{table}

\section{Further Details of Each Plot in the Main Text}
Figure~\ref{fig:case1} and Figure~\ref{fig:case2} used the sliding window method to average the adjacent $\lambda=50$ points into one point to make the curve smoother and facilitate analysis.

\section{Dataset Details}

We present the train-test partitioning and max generation length of the dataset used for evaluation in the article in Table~\ref{tab:dataset_detail}.
\begin{table}[h]
\centering
\begin{tabular}{c|ccccc|cc} 
\midrule[0.15ex]
Dataset  & Train/Test & Max Generation Length\\
\midrule[0.05ex]
GSM8K  & 7.47K/1.32K & 2048 \\
MATH  & 5.74K/0.5K & 2048 \\
KK  & 4.50K/0.5K & 2048 \\
\midrule[0.15ex]
\end{tabular}
\caption{\textbf{Dataset Details} \label{tab:dataset_detail}}
\end{table}

\section{Example Evaluation}
This appendix provides example evaluations of \sysname on the GSM8K dataset, as shown in Table~\ref{tab:exp}.

\begin{table}[htb]\centering
\raggedright{\textbf{Example Evaluation from \sysname}}

\resizebox{\linewidth}{!}{
\begin{tabular}{p{1.1\linewidth}} 
\midrule[0.3ex]

\#\#\# GSM8K Dataset: \\
\textbf{Question:} Eliza's rate per hour for the first 40 hours she works each week is \$10. She also receives an overtime pay of 1.2 times her regular hourly rate. If Eliza worked for 45 hours this week, how much are her earnings for this week?

\textbf{Answer:} Eliza is entitled to 45 - 40 = 45-40=5 hours overtime pay. Her hourly rate for the overtime pay is \$10 $\times$ 1.2 = \$10*1.2=12. So, Eliza will receive 12 $\times$ 5 = \$12*5=60 for overtime pay. Her regular weekly earning is \$10 $\times$ 40 = \$10*40=400. Thus, Eliza will receive a total of \$400 + \$60 = \$400+60=460 for this week's work. \#\#\# 460

\#\#\# Output from \sysname: \\
\textbf{Question:} Eliza's rate per hour for the first 40 hours she works each week is \$10. She also receives an overtime pay of 1.2 times her regular hourly rate. If Eliza worked for 45 hours this week, how much are her earnings for this week?

\textbf{Completion:} Okay, let's see. The problem is about Eliza's earnings for the week. Hmm, so she works 45 hours this week. The first 40 hours she gets \$10 per hour. Then there's overtime pay, which is 1.2 times her regular rate. Wait, so overtime is for the hours beyond 40, right? So she worked 45 hours total. That means 40 regular hours and 5 overtime hours. First, I need to calculate her earnings for the regular hours. So 40 hours times \$10 per hour. Let me do that: 40 * 10 = \$400. That parts seems straightforward. Now, the overtime pay. Her regular rate is \$10, so overtime is 1.2 times that. Let me calculate the overtime rate first. 1.2 * 10 = \$12 per hour. So each overtime hour she earns \$12. She worked 5 overtime hours. So 5 * 12 = \$60. Then, adding the regular and overtime earnings together: \$400 + \$60 = \$460. So her total earnings for the week should be \$460. Let me check if I did that right. Wait, 40 hours at \$10 definitely \$400. Then 5 hours at \$12 is \$60. Yep, adding them gives \$460. That seems right. I don't think I missed anything here. The problem didn't mention any other deductions or taxes, so I guess that's the total. Yeah, I think that's correct. Eliza's total earnings for the week can be calculated by breaking down her hours into regular and overtime pay: 
1. "Regular Hours Earnings": 40 hours $\times$ \$10/hour = \$400, 
2. "Overtime Hours Earnings": Overtime rate: 1.2 $\times$ \$10 = \$12/hour, Overtime hours: 45 - 40 = 5 hours, Total overtime earnings: \$12 $\times$ 5 = \$60, 
3. "Total Earnings": \$400 + \$60 = \$460, Answer: \boxed{460} The answer is 460!

\\ \bottomrule[0.3ex]
\end{tabular}
}
\caption{Example Evaluation from \sysname}
\label{tab:exp}
\end{table}

\end{document}